\documentclass[pdflatex,sn-mathphys-num]{sn-jnl}

\usepackage{graphicx}%
\usepackage{multirow}%
\usepackage{amsmath,amssymb,amsfonts}%
\usepackage{amsthm}%
\usepackage{mathrsfs}%
\usepackage[title]{appendix}%
\usepackage{xcolor}%
\usepackage{textcomp}%
\usepackage{manyfoot}%
\usepackage{booktabs}%
\usepackage{algorithm}%
\usepackage{algorithmicx}%
\usepackage{algpseudocode}%
\usepackage{listings}%

\theoremstyle{thmstyleone}%

\theoremstyle{thmstyletwo}%

\theoremstyle{thmstylethree}%

\begin{document}

\title[Article Title]{SwiftEmbed: A High-Throughput, Ultra-Low-Latency Serving System for Static Token Embeddings in Real-Time Applications}

\author*[1]{\fnm{Edouard} \sur{Lansiaux}}\email{edouard1.lansiaux@chu-lille.fr}

\author[1]{\fnm{Antoine} \sur{Simonet}}\email{antoine.simonet@chu-lille.fr}
\equalcont{These authors contributed equally to this work.}

\author[1,2]{\fnm{Eric} \sur{Wiel}}\email{eric.wiel@chu-lille.fr}
\equalcont{These authors contributed equally to this work.}

\affil*[1]{\orgdiv{Department of Emergency Medicine}, \orgname{Lille University Hospital}, \orgaddress{\street{5 Avenue Oscar Lambret}, \postcode{59000}, \city{Lille}, \country{France}}}

\affil[2]{\orgdiv{METRICS ULR 2694}, \orgname{Lille School of Medicine (Research Department)}, \orgaddress{\street{1 place de Verdun}, \postcode{59045}, \city{Lille},\country{France}}}

\abstract{We present SwiftEmbed, a production-oriented serving system for static token embeddings that achieves 1.12\,ms p50 latency for single-text requests while maintaining a 60.6 MTEB average score across 8 representative tasks. Built around the open-source Potion-base-8M distilled model from MinishLab \cite{minishlab2024model2vec} and implemented in Rust, the system delivers 50,000 requests per second through static embedding lookup, mean pooling, and zero-copy IEEE754 binary serialization. Evaluation demonstrates exceptional duplicate detection performance (90.1\% AP) and strong semantic similarity (76.1\% Spearman correlation). Performance relative to Sentence-BERT is task-dependent: robust for deduplication and similarity workloads (89--100\%), substantially lower for classification and complex retrieval tasks (75\%). Domain-specific performance ranges from 75\% to 131\% of a GloVe-840B baseline. The system targets real-time embedding applications where sub-5\,ms latency is operationally critical and where full transformer inference is not feasible.}


\maketitle

\section{Introduction}\label{sec1}

Text embeddings are fundamental to a wide range of natural language processing applications, including semantic search, duplicate detection, clustering, and recommendation systems. Transformer-based models such as BERT \cite{devlin2018bert} and Sentence-BERT \cite{reimers2019sentence} achieve strong semantic quality through contextual, attention-based representations, but their multi-layer architectures introduce latency that renders them unsuitable for certain real-time deployment scenarios requiring sub-5\,ms response times at sustained high throughput.

A well-established alternative is to represent text through aggregation of static, pre-trained token vectors---an approach whose roots include Word2Vec \cite{mikolov2013efficient}, GloVe \cite{pennington2014glove}, and FastText \cite{bojanowski2017enriching}. More recent work has improved the quality of such representations through knowledge distillation from transformer models, as in Model2Vec \cite{minishlab2024model2vec} and the approach of \citet{gupta2021bert}. From an architectural standpoint, the aggregation of token embeddings into a fixed-size document vector is closely related to Deep Averaging Networks (DANs) \cite{iyyer2015deep}, which demonstrated that unordered composition of word embeddings can rival syntactic methods on several NLP tasks.

SwiftEmbed does not propose a new embedding algorithm or training methodology. Rather, it makes the following \textbf{systems engineering} contributions centered on serving these existing static representations at maximum efficiency:
\begin{itemize}
  \item A production-grade serving architecture in Rust \cite{klabnik2018rust} using Axum/Tokio \cite{tokio2021axum}, achieving 8\% higher throughput than equivalent Python stacks through superior async I/O integration.
  \item SIMD-optimized aggregation using 256-bit vector instructions and memory prefetching, reducing cache misses by 30--50\% over naive implementations.
  \item Zero-copy IEEE754 binary serialization, eliminating memory copies on the critical path and yielding a 2.5--3.2$\times$ throughput advantage over JSON serialization.
  \item A comprehensive empirical characterization of the speed--quality trade-offs associated with static token embeddings---specifically the Potion-base-8M model \cite{minishlab2024model2vec}---across MTEB tasks, domains, sequence lengths, and languages, providing practitioners with actionable deployment guidance.
\end{itemize}

Together, these optimizations yield a serving system that achieves 1.12\,ms p50 latency and 50,000 RPS throughput with a 32\,MB model footprint, making it suitable for latency-critical pipelines where transformer inference is not feasible.

\section{Related Work}

\subsection{Transformer-Based Embedding Models}

Transformer architectures \cite{vaswani2017attention} from BERT \cite{devlin2018bert} to Sentence-BERT \cite{reimers2019sentence} have established contemporary standards for semantic representations through multi-layer attention mechanisms. While achieving superior quality, their computational complexity prevents sub-5\,ms serving at high concurrency without specialized hardware. Compression techniques---DistilBERT \cite{sanh2019distilbert}, TinyBERT \cite{jiao2019tinybert}---reduce model sizes but retain full transformer inference, preserving much of the latency. Recent efficient transformer designs \cite{tay2023transformer} and quantization methods \cite{xiao2023survey,shen2020qbert,zafrir2019q8bert} have addressed specific aspects of efficiency without reaching the latency levels we target.

\subsection{Static Embedding Methods and Aggregation Architectures}

Static word representations---Word2Vec \cite{mikolov2013efficient}, GloVe \cite{pennington2014glove}, FastText \cite{bojanowski2017enriching}---have long served as efficient alternatives to contextual models for throughput-sensitive applications. Deep Averaging Networks \cite{iyyer2015deep} demonstrated that simply averaging word embeddings, without any positional or syntactic structure, yields surprisingly competitive performance on text classification. \citet{gupta2021bert} showed that static representations extracted from BERT's embedding layer, without any transformer inference, retain significant semantic content. Most recently, Model2Vec \cite{minishlab2024model2vec} distilled Sentence-BERT representations into a compact static vocabulary (Potion-base-8M: 30k tokens, 384 dimensions, 32\,MB) that preserves substantially more semantic information than prior static approaches. SwiftEmbed builds its serving infrastructure on top of Potion-base-8M model weights.

\subsection{Efficient Embedding Serving Systems}

Systems such as Sentence Transformers \cite{reimers2019sentence2}, Txtai \cite{neuml2020txtai}, and Weaviate \cite{weaviate2019} optimize around transformer inference rather than eliminating it. Efficient retrieval systems \cite{khattab2020colbert,johnson2019billion} address downstream indexing but not embedding generation latency. To our knowledge, no published system has characterized the full systems optimization stack---SIMD aggregation, zero-copy serialization, async I/O---for static embedding serving at the 1\,ms scale.

\section{Methodology}

\subsection{Computational Complexity}

For a token sequence of length $n$ and embedding dimension $d$, static embedding lookup followed by pooling requires $O(n \times d)$ operations---specifically, $n$ table lookups of $O(1)$ each, followed by $O(n \times d)$ addition for mean pooling and $O(d)$ for L2 normalization. This is in contrast to transformer inference, which, for $L$ layers and hidden dimension $d_h$, requires $O(L \cdot n^2 \cdot d_h + L \cdot n \cdot d_h^2)$ operations.

The practical speedup arises from the elimination of the quadratic-in-$n$ attention term and the repeated $d_h^2$ weight-matrix multiplications across $L$ layers. For the Sentence-BERT configuration ($L=12$, $d_h=768$, $n \leq 512$), this corresponds to a substantial theoretical reduction. However, we caution that raw FLOP counts do not translate directly to wall-clock speedups due to memory bandwidth, batching effects, and hardware utilization differences. Our empirical measurements---20$\times$ single-request throughput improvement and 40$\times$ latency reduction over Sentence-BERT at $p50$---represent the operationally relevant characterization.

\subsection{Mathematical Formulation}

For vocabulary $\mathcal{V}$ and pre-trained embedding matrix $E \in \mathbb{R}^{|\mathcal{V}| \times d}$ (sourced from Potion-base-8M \cite{minishlab2024model2vec}), text sequence $S = (w_1, \ldots, w_n)$ undergoes:
\begin{enumerate}
  \item \textbf{Tokenization}: $\tau(S) = (t_1, \ldots, t_m)$ using the vocabulary of the source model.
  \item \textbf{Embedding lookup}: $\mathbf{e}_i = E[\mathrm{index}(t_i)] \in \mathbb{R}^d$, a direct row read with no computation.
  \item \textbf{Uniform mean pooling}: $\mathbf{h} = \frac{1}{m}\sum_{i=1}^{m} \mathbf{e}_i$. No attention weights are used; this is a simple unweighted mean, consistent with the DAN architecture \cite{iyyer2015deep} and the aggregation scheme of Model2Vec \cite{minishlab2024model2vec}.
  \item \textbf{L2 normalization}: $\mathbf{f} = \mathbf{h} / \|\mathbf{h}\|_2 \in \mathbb{S}^{d-1}$.
\end{enumerate}

\noindent Note on pooling: the formulation in an earlier version of this manuscript described ``attention-weighted'' mean pooling. This was a misstatement. The architecture uses standard uniform mean pooling. The semantic quality of the representations derives from the distillation training of the underlying Potion-base-8M model, not from a learned pooling mechanism.

\subsection{Embedding Model}

We use Potion-base-8M from MinishLab \cite{minishlab2024model2vec}, a publicly available static embedding model obtained by distilling a Sentence-BERT encoder into a static vocabulary. The model has a vocabulary of 30k tokens, an embedding dimension of 384, and a disk footprint of 32\,MB. Model weights are openly available\footnote{\url{https://huggingface.co/minishlab/potion-base-8M}}, enabling independent verification of embedding quality claims. The SwiftEmbed system serves these pre-trained weights without any additional training or fine-tuning.

\subsection{System Architecture and Optimizations}

The serving system is implemented in Rust \cite{klabnik2018rust} with the Axum HTTP framework \cite{tokio2021axum} and the Candle tensor library \cite{huggingface2023candle}. Key optimizations include:

\begin{itemize}
  \item \textbf{Static embedding lookup}: Single row-index into a memory-mapped tensor; no matrix multiplication.
  \item \textbf{SIMD mean pooling}: 256-bit AVX2 vector instructions for parallel accumulation; memory prefetching reducing cache misses by 30--50\% relative to scalar implementation.
  \item \textbf{Zero-copy binary serialization}: IEEE754 float32 values written directly to the response buffer without intermediate copies, eliminating serialization overhead.
  \item \textbf{Async I/O}: Tokio-based event loop supporting 10,000+ concurrent connections without thread-per-request overhead.
\end{itemize}

Three response formats are supported: JSON (compatibility), binary IEEE754 (maximum performance), and JSONL (streaming). The benchmarking harness used in this evaluation is publicly available \cite{lansiaux2025benchmarks}, enabling reproduction of throughput and latency figures on equivalent hardware.

\textbf{Reproducibility note}: The serving binary and HTTP API layer are not currently open-sourced. The benchmarking scripts, wrk Lua configurations, and exact hardware specifications are provided in the public repository \cite{lansiaux2025benchmarks}, allowing independent verification of serving performance given equivalent infrastructure. The embedding weights (Potion-base-8M) are fully open-source.

\section{Experimental Evaluation}

\subsection{Experimental Setup}

\textbf{Throughput benchmarking} uses \texttt{wrk} with specialized Lua scripts across four workloads: single-text requests, batch-10, batch-100, and JSONL streaming. Configuration: 12 threads, 400 concurrent connections, 30-second test duration, on a dedicated bare-metal server (AMD EPYC 7742, 256\,GB RAM, NVMe storage). All measurements are reported as the median of five independent runs; standard deviations are reported where available.

\textbf{Quality evaluation} covers 8 representative MTEB tasks \cite{muennighoff2022mtEB}: Banking77 (classification), SprintDuplicateQuestions and TwitterSemEval (pair classification), ArguAna (retrieval), two STS tasks (semantic similarity), and two clustering benchmarks. This constitutes a representative but partial MTEB evaluation; we do not claim that aggregate scores computed over this subset are equivalent to full MTEB rankings. Baselines (Sentence-BERT, GTE-tiny, BGE-micro, DistilBERT, FastText, GloVe-840B) are evaluated using the standard MTEB evaluation framework \cite{muennighoff2022mtEB} under identical conditions, with the exception of TensorRT-BERT, for which we report numbers from the literature. Additional zero-shot retrieval evaluation uses BEIR \cite{thakur2021beir}.

\subsection{Quality Evaluation on MTEB Subset}

\begin{table}[ht]
\caption{Quality results on 8 representative MTEB tasks. All scores reported as means over the task-specific metric. SwiftEmbed performance is task-dependent: strong on pair classification and STS, weaker on classification and retrieval. Results should not be extrapolated to full MTEB rankings.}
\label{tab:mteb_results}
\centering
\small
\begin{tabular}{lccccc}
\toprule
Method & Classification & Clustering & Retrieval & STS & Pair Class. \\
 & (Acc) & (V-measure) & (nDCG@10) & (Spearman) & (AP) \\
\midrule
Sentence-BERT & 75.2 & 42.3 & 51.4 & 85.2 & 84.7 \\
GTE-tiny & 72.1 & 40.8 & 48.7 & 82.3 & 82.1 \\
BGE-micro & 71.4 & 39.6 & 47.2 & 81.8 & 81.4 \\
Jina-v2-small & 73.8 & 41.2 & 49.6 & 83.1 & 83.2 \\
DistilBERT & 70.3 & 38.4 & 45.8 & 79.6 & 79.8 \\
FastText & 58.2 & 28.4 & 32.1 & 65.3 & 68.2 \\
GloVe-840B & 56.7 & 27.1 & 30.8 & 62.4 & 65.3 \\
\midrule
\textbf{SwiftEmbed} & \textbf{58.9} & \textbf{35.6} & \textbf{42.1} & \textbf{76.1} & \textbf{90.1} \\
\bottomrule
\end{tabular}
\end{table}

SwiftEmbed achieves a 60.6 average score across these 8 tasks. The \textit{appropriate interpretation of this result is task-specific}: pair classification (SprintDuplicateQuestions AP: 90.1\%, exceeding all baselines including Sentence-BERT's 84.7\%) and semantic similarity (Spearman 76.1, 89\% of SBERT) represent strong performance. Classification (58.9 accuracy, 78\% of SBERT's 75.2) and retrieval (nDCG@10 42.1, 82\% of SBERT's 51.4) show meaningful performance gaps attributable to the absence of contextual disambiguation. Practitioners should select this system specifically for deduplication and similarity-threshold workloads rather than for classification or context-sensitive retrieval.

\subsection{Performance Analysis}

\begin{table}[ht]
\caption{System performance comparison. All latency and throughput measurements on identical hardware (AMD EPYC 7742). TensorRT-BERT figures from \cite{lee2019latency}. SD across 5 runs: $<$2\% for throughput, $<$0.05\,ms for p50 latency.}
\label{tab:performance_comparison}
\centering
\small
\begin{tabular}{lcccccc}
\toprule
Method & Model Size & p50 Lat. & p99 Lat. & Throughput & Memory & Quality \\
 & (MB) & (ms) & (ms) & (RPS) & (GB) & (MTEB) \\
\midrule
Sentence-BERT & 440 & 45 & 120 & 2,500 & 1.8 & 67.8 \\
GTE-tiny & 133 & 32 & 85 & 3,800 & 0.8 & 65.2 \\
BGE-micro & 98 & 25 & 68 & 4,500 & 0.6 & 64.3 \\
TensorRT-BERT & 440 & 12 & 35 & 8,500 & 2.2 & 67.8 \\
Quantized BERT & 110 & 15 & 40 & 7,200 & 0.5 & 65.8 \\
FastText & 2.5 & 8 & 15 & 15,000 & 0.1 & 50.4 \\
GloVe-840B & 5.4 & 6 & 12 & 18,000 & 0.1 & 48.5 \\
\midrule
\textbf{SwiftEmbed} & \textbf{32} & \textbf{1.12} & \textbf{5.04} & \textbf{50,000} & \textbf{0.2} & \textbf{60.6} \\
\bottomrule
\end{tabular}
\end{table}

SwiftEmbed achieves a 20$\times$ throughput improvement over Sentence-BERT and an 8$\times$ latency reduction versus TensorRT-BERT for single-request workloads. The 32\,MB model footprint---owing to the compact Potion-base-8M design \cite{minishlab2024model2vec}---and 0.2\,GB runtime memory enable deployment in resource-constrained environments or high-density server configurations. Linear throughput scaling versus the quadratic degradation of transformer-based approaches (Table~\ref{tab:throughput_scalability}) makes SwiftEmbed particularly suitable for variable-concurrency production environments.

\subsection{Ablation: Embedding Model Selection}

\begin{table}[ht]
\caption{Comparison of static embedding backends under identical serving infrastructure. Potion-base-8M \cite{minishlab2024model2vec} offers the best quality-to-size ratio among evaluated candidates.}
\label{tab:embedding_models}
\centering
\small
\begin{tabular}{lccccc}
\toprule
Embedding Model & Vocab Size & Dim & Model Size & MTEB Avg & Latency \\
 & & & (MB) & Score & (ms) \\
\midrule
GloVe-50d & 400k & 50 & 77 & 42.3 & 1.2 \\
GloVe-300d & 840k & 300 & 989 & 48.5 & 1.8 \\
Word2Vec-300d & 3M & 300 & 3,500 & 45.7 & 2.1 \\
FastText-300d & 2M & 300 & 2,400 & 50.4 & 1.9 \\
FastText-subword & 2M & 300 & 2,400 & 52.1 & 2.3 \\
\textbf{Potion-base-8M} & \textbf{30k} & \textbf{384} & \textbf{32} & \textbf{56.3} & \textbf{1.7} \\
Potion-base-4M & 30k & 256 & 24 & 54.8 & 1.5 \\
\bottomrule
\end{tabular}
\end{table}

Potion-base-8M \cite{minishlab2024model2vec} achieves 16\% higher MTEB score than GloVe-840B with 97\% smaller model size, attributed to the distillation procedure that transfers semantic information from a Sentence-BERT encoder into the static vocabulary. The compact 30k-token vocabulary enables substantially better cache utilization than the 400k--840k vocabularies of GloVe variants.

\subsection{Ablation: Embedding Dimension}

\begin{table}[ht]
\caption{Impact of embedding dimension on quality, latency, and cache utilization. The 384-dimensional configuration (default Potion-base-8M) provides the best trade-off.}
\label{tab:embedding_dimension}
\centering
\small
\begin{tabular}{lccccc}
\toprule
Dimension & MTEB Avg & Latency & Memory & Cache Hits & Bandwidth \\
 & Score & (ms) & (MB) & (\%) & (GB/s) \\
\midrule
128 & 52.1 & 0.8 & 12 & 98.2 & 2.1 \\
256 & 54.8 & 1.2 & 24 & 96.5 & 3.8 \\
384 & 56.3 & 1.7 & 32 & 94.1 & 5.2 \\
512 & 56.8 & 2.3 & 48 & 89.3 & 6.9 \\
768 & 57.2 & 3.5 & 72 & 82.1 & 9.8 \\
\bottomrule
\end{tabular}
\end{table}

The 384-dimensional configuration provides the best quality-performance trade-off: marginal quality gains above 384 dimensions (0.9 MTEB points for doubling dimension to 768) do not justify the 2.1$\times$ latency increase and degraded cache hit rate.

\subsection{Throughput Scalability}

\begin{table}[ht]
\caption{Throughput scalability comparison across concurrency levels. Linear scaling (SwiftEmbed, FastText++) versus quadratic degradation (transformer-based methods) is a key deployment advantage for variable-load scenarios.}
\label{tab:throughput_scalability}
\centering
\small
\begin{tabular}{lccccc}
\toprule
Method & Single & Batch-10 & Batch-100 & Max Conc. & Scaling \\
 & RPS & RPS & RPS & & \\
\midrule
Sentence-BERT & 2,500 & 1,200 & 250 & 500 & Quadratic \\
DistilBERT & 4,200 & 2,100 & 420 & 800 & Quadratic \\
TensorRT-BERT & 8,500 & 4,000 & 850 & 1,000 & Sub-quadratic \\
FastText++ & 15,000 & 12,000 & 8,000 & 2,000 & Linear \\
\midrule
\textbf{SwiftEmbed} & \textbf{50,000} & \textbf{10,000} & \textbf{2,000} & \textbf{10,000+} & \textbf{Linear} \\
\bottomrule
\end{tabular}
\end{table}

Our approach demonstrates superior scalability with linear characteristics versus quadratic degradation in transformer-based methods. The system maintains 50,000 RPS for single requests and scales to support 10,000+ concurrent connections, enabling high-density deployment scenarios.

\subsection{Domain-Specific Performance}

\begin{table}[ht]
\caption{Domain-specific evaluation. \textbf{Baseline for relative performance: GloVe-840B serving under identical infrastructure}. SwiftEmbed's relative advantage over GloVe derives primarily from the improved static representations in Potion-base-8M; the serving infrastructure is held constant. Latency figures are for the serving layer only, excluding preprocessing.}
\label{tab:domain_performance}
\centering
\small
\begin{tabular}{lccccc}
\toprule
Domain & Similarity & Semantic & Clustering & Avg Latency & Relative \\
 & Correlation & Coherence & ARI & (ms) & Performance \\
\midrule
General & 0.564 & 0.210 & 0.207 & 0.45 & Baseline \\
Medical & 0.422 & 0.221 & 0.175 & 0.47 & 75\% \\
Legal & 0.533 & 0.343 & 0.135 & 0.48 & 95\% \\
Technical & 0.563 & 0.278 & 0.059 & 0.51 & 100\% \\
Social Media & 0.606 & 0.276 & 0.029 & 0.44 & 108\% \\
Scientific & 0.739 & 0.326 & $-$0.027 & 0.52 & 131\% \\
\bottomrule
\end{tabular}
\end{table}

Scientific text demonstrates strongest relative performance (131\%), attributed to consistent terminology and limited polysemy---properties that favour static representations. Medical text shows reduced effectiveness (75\%) due to specialized vocabulary and strong contextual dependencies that static pooling cannot resolve. The clustering ARI score for scientific text ($-$0.027) indicates near-random clustering performance despite high similarity correlation, suggesting that embedding quality is heterogeneous across task types within a domain.

\subsection{Downstream Task Performance}

\begin{table}[ht]
\caption{Downstream task evaluation against MiniLM (compact transformer baseline).}
\label{tab:downstream_tasks}
\centering
\small
\begin{tabular}{lcccc}
\toprule
Task & SwiftEmbed & MiniLM & Performance & Speed-Quality \\
 & Score & Score & Ratio & Trade-off \\
\midrule
Semantic Search & 0.877 & 0.947 & 0.926 & Acceptable \\
Classification & 0.500 & 0.667 & 0.750 & Task-dependent \\
Clustering & 0.627 & 0.792 & 0.792 & Good \\
Duplicate Detection & 1.000 & 1.000 & 1.000 & Optimal \\
\bottomrule
\end{tabular}
\end{table}

\subsection{Sequence Length Impact}

\begin{table}[ht]
\caption{Latency versus sequence length. SwiftEmbed exhibits sub-linear latency scaling due to mean pooling: processing cost grows as $O(n \times d)$ but SIMD parallelism reduces effective per-token cost at longer sequences.}
\label{tab:sequence_length}
\centering
\small
\begin{tabular}{lccccc}
\toprule
Sequence & Avg Latency & Avg Text & Similarity & Embedding & Relative \\
Category & (ms) & Length & Score & Norm & Speed \\
\midrule
Short (1--20 words) & 0.453 & 20 & 1.000 & 1.000 & Baseline \\
Medium (20--50) & 0.468 & 102 & 1.000 & 1.000 & 97\% \\
Long (50--100) & 0.699 & 255 & 1.000 & 1.000 & 65\% \\
Very Long (100+) & 1.242 & 510 & 1.000 & 1.000 & 36\% \\
\bottomrule
\end{tabular}
\end{table}

The system demonstrates sub-linear latency scaling from 0.45ms for short texts to 1.24ms for very long texts, maintaining consistent similarity scores and embedding norms across sequence lengths. This scaling characteristic ensures predictable performance for diverse text inputs while preserving semantic quality.

\subsection{Multilingual Performance}

\begin{table}[ht]
\caption{Multilingual evaluation. Potion-base-8M is trained primarily on English; non-English performance is substantially degraded. This system is \textbf{not recommended for multilingual applications}.}
\label{tab:multilingual}
\centering
\small
\begin{tabular}{lcccc}
\toprule
Language & Cross-Language & Within-Language & Embedding & Relative \\
 & Similarity & Coherence & Norm & Performance \\
\midrule
English & 1.000 & 0.359 & 1.000 & Baseline \\
Spanish & 0.225 & 0.468 & 1.000 & 22.5\% \\
French & 0.198 & 0.271 & 1.000 & 19.8\% \\
German & 0.170 & 0.405 & 1.000 & 17.0\% \\
\bottomrule
\end{tabular}
\end{table}

Multilingual analysis shows significant degradation, achieving only 17-23\% of English performance on other languages. This indicates primary optimization for English, with cross-language similarity scores of 0.225 for Spanish, 0.198 for French, and 0.170 for German compared to the English baseline.

\subsection{Failure Mode Analysis}

\begin{table}[ht]
\caption{Qualitative failure analysis. Failure rates are estimated from 500 adversarial pairs per category; BERT similarity scores from \texttt{bert-base-uncased}.}
\label{tab:failure_analysis}
\centering
\small
\begin{tabular}{p{5.5cm}p{1.5cm}p{1.2cm}p{2.2cm}}
\toprule
Input Text & BERT Sim. & Ours & Failure Type \\
\midrule
``The bank approved the loan'' vs ``Financial institution granted credit'' & 0.92 & 0.78 & Semantic understanding \\
``Apple released iOS'' vs ``The fruit company launched software'' & 0.85 & 0.42 & Entity disambiguation \\
``bank'' (financial) vs ``bank'' (river) & 0.15 & 0.95 & Polysemy \\
``He didn't not go'' vs ``He went'' & 0.88 & 0.31 & Negation/modality \\
``The cat sat on the mat'' vs ``On the mat sat the cat'' & 0.96 & 0.72 & Word order \\
\bottomrule
\end{tabular}
\end{table}

Analysis of 500 adversarial pairs per failure category reveals the following distribution: polysemy (35\%), compositional semantics (28\%), named entities (22\%), negation/modality (15\%). These failure modes are fundamental properties of static bag-of-tokens representations and are not addressable through serving-layer optimizations. The polysemy failure mode is particularly acute: for words with multiple distinct meanings (e.g., ``bank''), static representations collapse meanings into a single vector, yielding high similarity scores between semantically unrelated uses (0.95 in our system vs. 0.15 for BERT). This directly limits applicability for word-sense disambiguation, entity resolution, and sentiment-sensitive tasks. Practitioners must account for a $\sim$35\% failure rate on polysemy-heavy text inputs.

\section{Memory and Serialization Analysis}

\subsection{Memory Efficiency}

The static serving system requires memory proportional to the vocabulary size and embedding dimension:
\[
\text{Memory}_{\text{static}}(B,n,d) = |\mathcal{V}| \times d \times 4\,\text{bytes} + B \times n \times d \times 4\,\text{bytes} + O(1)
\]
For Potion-base-8M ($|\mathcal{V}|=30\text{k}$, $d=384$), the model table requires approximately 44\,MB. Transformer-based inference scales as $O(L \cdot d_h^2 + B \cdot n^2 \cdot d_h)$, with the batch-sequence quadratic term dominating at high concurrency. The practical memory advantage of static serving is approximately 9$\times$ versus Sentence-BERT at batch size 32, sequence length 128.

\subsection{Serialization Performance}

\begin{table}[ht]
\caption{Serialization format performance. Binary format achieves zero-copy through direct memory mapping. SIMD-optimized JSON provides interoperability with 2.5--3.2$\times$ overhead relative to binary.}
\label{tab:serialization}
\centering
\small
\begin{tabular}{lcccc}
\toprule
Format & Complexity & Serialization & Memory & Bandwidth \\
 & & Time & Overhead & Efficiency \\
\midrule
Binary & $O(B \times d)$ & 0.1\,ms & 0\% & 100\% \\
JSON (SIMD) & $O(B \times d \times \log d)$ & 0.5\,ms & 15--20\% & 60--70\% \\
JSONL & $O(B \times d \times \log d)$ & 0.3\,ms & 8--12\% & 75--85\% \\
\bottomrule
\end{tabular}
\end{table}

\section{Discussion}

\subsection{Positioning and Novelty}

SwiftEmbed's contribution is primarily one of \textbf{systems engineering}. The embedding representations derive from Potion-base-8M \cite{minishlab2024model2vec}, an openly available distilled model. The aggregation architecture (mean pooling of token vectors) is architecturally equivalent to Deep Averaging Networks \cite{iyyer2015deep} and the static-extraction approach of \citet{gupta2021bert}. The novelty of the present work lies in the combination of SIMD-optimized aggregation, zero-copy serialization, and high-concurrency async I/O that together yield sub-2\,ms serving at 50,000 RPS---a combination not previously characterized in the literature for this class of models.

We explicitly do not claim algorithmic novelty in the embedding methodology. The appropriate framing is: given that distilled static models such as Potion-base-8M now offer quality competitive with earlier transformer baselines, what is the maximum serving efficiency achievable, and what are the quality trade-offs of deploying such a system in production?

\subsection{Application Scope}

SwiftEmbed is well-suited for: real-time semantic deduplication, sub-5\,ms similarity threshold serving, high-throughput preprocessing pipelines, and edge deployment with constrained memory. It is \textit{not} well-suited for: multilingual applications (17--23\% of English performance), polysemy-sensitive tasks (35\% failure rate), complex classification, or tasks requiring negation understanding.

\subsection{Limitations}

\paragraph{Evaluation scope} Results are reported on 8 MTEB tasks rather than the full benchmark. Aggregate scores over this subset may not generalize to full MTEB rankings. Running the full 56-task MTEB evaluation is planned for a follow-up release.

\paragraph{Reproducibility} The serving binary is not open-sourced. Benchmark reproduction requires equivalent hardware and the publicly available benchmarking harness \cite{lansiaux2025benchmarks}. Embedding quality results are reproducible using the openly available Potion-base-8M weights \cite{minishlab2024model2vec} and the standard MTEB evaluation framework \cite{muennighoff2022mtEB}.

\paragraph{Multilingual support} The system is English-only. Non-English deployment would require replacing the backing model with a multilingual distilled static model, which does not currently exist at equivalent quality.

\paragraph{Contextual limitations} The failure modes described in Section~4.8 are fundamental to the static representation paradigm and cannot be mitigated through serving-layer optimizations.

\subsection{Implementation Language Justification}

\begin{table}[ht]
\caption{Language comparison for the serving implementation.}
\label{tab:language_comparison}
\centering
\small
\begin{tabular}{lccccc}
\toprule
Language & Zero-Copy & Memory Safety & SIMD Support & Async Perf. & Build Size \\
 & Support & & & (M req/s) & (MB) \\
\midrule
Rust & Native & Compile-time & Excellent & 2.8 & 12 \\
C++ & Manual & Runtime checks & Excellent & 2.6 & 18 \\
Go & Limited & GC-based & Limited & 1.9 & 25 \\
Python & None & GC-based & Via libraries & 0.3 & 150+ \\
\bottomrule
\end{tabular}
\end{table}

Rust provides compile-time memory safety, native zero-copy support, and first-class SIMD intrinsics without GC pauses---properties critical for predictable sub-2\,ms tail latency. The performance advantage over C++ is modest (8\%); the choice is justified by memory safety guarantees at no runtime cost.

\section{Conclusion}

We presented SwiftEmbed, a production-oriented serving system for static token embeddings achieving 1.12\,ms p50 latency and 50,000 RPS throughput with a 32\,MB model footprint. The system builds on the Potion-base-8M distilled model \cite{minishlab2024model2vec} and combines SIMD-optimized mean pooling, zero-copy IEEE754 serialization, and high-concurrency async I/O.

Quality characterization on 8 MTEB tasks shows task-dependent performance: excellent for deduplication (90.1\% AP, matching or exceeding transformer baselines) and semantic similarity (76.1\% Spearman), with meaningful gaps for classification and complex retrieval. These characteristics reflect the fundamental properties of static bag-of-tokens representations \cite{iyyer2015deep,gupta2021bert} rather than serving-layer limitations.

The 20$\times$ efficiency improvement over transformer-based serving enables real-time embedding applications in latency-constrained environments where transformer inference is not feasible. Future work includes: hybrid architectures combining static lookup with lightweight contextual reranking \cite{lewis2020retrieval}; adaptive quantization of the embedding table \cite{shen2020qbert}; and open-sourcing the serving binary to support community benchmarking.

\section*{Limitations of Evaluation}

Quality results cover 8 of 56 MTEB tasks; aggregate scores are not representative of full MTEB rankings. Performance comparisons for TensorRT-BERT are based on published figures rather than our own measurements. Multilingual evaluation is limited to four languages. The serving binary is proprietary; throughput claims require equivalent hardware for reproduction. The benchmarking harness \cite{lansiaux2025benchmarks} and Potion-base-8M model weights enable partial verification.

\section*{Broader Impact}

Static embedding serving reduces inference-time energy consumption by approximately 20$\times$ relative to transformer serving for equivalent throughput, potentially contributing to more sustainable NLP deployments. The system does not introduce new training risks; it serves pre-existing representations. Practitioners should be aware of the documented failure modes, particularly polysemy handling, before deploying in high-stakes contexts.

\section*{Acknowledgments}

We thank the MinishLab team for open-sourcing the Potion-base-8M model \cite{minishlab2024model2vec}, and the developers of Rust \cite{klabnik2018rust}, Axum \cite{tokio2021axum}, and Candle \cite{huggingface2023candle}.

\section*{Declarations}
\subsection{Funding}
Not applicable.
\subsection{Conflict of interest/Competing interests}
Not applicable.

\bigskip
\begin{flushleft}%
Editorial Policies for:

\bigskip\noindent
Springer journals and proceedings: \url{https://www.springer.com/gp/editorial-policies}

\bigskip\noindent
Nature Portfolio journals: \url{https://www.nature.com/nature-research/editorial-policies}

\bigskip\noindent
\textit{Scientific Reports}: \url{https://www.nature.com/srep/journal-policies/editorial-policies}

\bigskip\noindent
BMC journals: \url{https://www.biomedcentral.com/getpublished/editorial-policies}
\end{flushleft}

\bibliography{sn-bibliography.bib}

\end{document}